# Causal Knowledge Extraction from Scholarly Papers in Social Sciences

Victor Zitian Chen[†], Felipe Montano-Campos[‡], Wlodek Zadrozny[*]

June 11th, 2020

**Abstract**

The scale and scope of scholarly articles today are overwhelming human researchers who seek to timely digest and synthesize knowledge. In this paper, we seek to develop natural language processing (NLP) models to accelerate the speed of extraction of relationships from scholarly papers in social sciences, identify hypotheses from these papers, and extract the cause-and-effect entities. Specifically, we develop models to 1) classify sentences in scholarly documents in business and management as hypotheses (hypothesis classification), 2) classify these hypotheses as causal relationships or not (causality classification), and, if they are causal, 3) extract the cause and effect entities from these hypotheses (entity extraction). We have achieved high performance for all the three tasks using different modeling techniques. Our approach may be generalizable to scholarly documents in a wide range of social sciences, as well as other types of textual materials.

Our code is available at: https://github.com/felipemontano10/NLP_CausalityExtraction

**Acknowledgements:** We owe thanks to Vanessa Alwan, John Barry, Marad Elsaify, Venetia Wong, and Haozhe Zhang for their constructive inputs to the early version of this research.

[†]Belk College of Business, University of North Carolina at Charlotte. Email:zchen23@uncc.edu.
[‡]Department of Computer Science, Duke University. Email: jose.montano.campos@duke.edu (Corresponding author).
[*]College of Computing and Informatics, University of North Carolina at Charlotte. Email:wzadrozn@uncc.edu.

# 1     Introduction

A research project in social sciences typically starts with a thorough review and synthesis of the literature for known principles about the relationships among established concepts, identifying the convergence (or lack thereof) of such relationships among identical concepts as well as new research agenda and untested hypotheses. However, the scale and scope of scholarly articles today are overwhelming human researchers who seek to timely digest and synthesize knowledge. Limited resources and professional time of researchers are often spent on tedious jobs of knowledge extraction, and yet these efforts are not sufficiently thorough.

In this paper, we seek to develop natural language processing (NLP) models to accelerate the speed of extraction of relationships from scholarly papers in social sciences, identify hypotheses from these papers, and extract the cause-and-effect entities. We focus on causal statements as they are meaningful for prescribing evidence-based managerial and policy interventions.

Our main research objective is to use machines to extract causal relationships from a large corpus of articles in social sciences. We focus on causal relationships that are deliberately developed conjectures as testable hypotheses, because these relationships form important scientific basis for prescribing managerial and policy interventions. For this paper, we are interested in detecting hypothesis statements, classifying casual statements, and extracting cause- and effect entities.

An association between two entities means that there exists some predictive relationship or connection between them. However, an association does not always imply causation, the latter of which is a stronger statement about an association (Pearl, 2009). As a result, the process of discovering the association and making causal inferences between two entities is at the core in all scientific disciplines. It is important for social sciences, where scholars, teachers, and consultants often draw empirical evidence to make prescriptions for managerial and policy practices.

Specifically, we first built an initial training dataset from 125 publications in social sciences. It is built based on an algorithm using keywords to extract causal and directional hypothesis statements, followed by manual identification of whether a statement is associative (e.g., A relates to B) or causal (e.g., if A increases, then B decreases) as well as annotations of the cause, the effect, and the direction (i.e., positive, negative, or nonlinear) of the relationship specified in each statement. After extracting the hypothesis sentences and manually classifying them as explained above, we then randomly selected a relatively identical sample size of non-hypothesis statements from the same publications.  Through the training set we reduced publications into a set of sentences that contain the information we need for developing our NLP models. For hypothesis statements, we labelled each of the extracted sentences with four features: the cause, the effect, the direction of the relationship, and whether this relationship is causal or not (causality). This practice simulates the process of information reduction through which researchers deal with a large literature. By reducing a large volume of publications into an annotated training dataset, researchers can simply analyze and organize the four features to succinctly understand the main findings of the literature.

After constructing our training dataset, we continued to perform our classification and extraction tasks. Our *first* task was **hypothesis classification** where we tried to identity if a sentence is a hypothesis or not. For this task we used the *fastText* model. *fastText* is an open-source library that does both word representations and text classification. This model has similar performance of accuracy as deep learning models but at a faster speed (Zolotov & Kung, 2017). Our *second* task was **causality classification**, where we classified sentences as whether they are causal statements or not. For this task, we found that logistic regression outperforms other methods. Furthermore, similar to prior works (Catalyst Team, 2016), we found that models using bag-of-words features outperformed those using other features. Lastly, our *third* task was **entity extraction**. Specifically, we extracted the cause and effect entities from causal statements. We used the LSTM model and found good overall performance.

We represent, to our knowledge, the first effort to develop causal knowledge extraction models for scholarly papers in social sciences. Thus, our major contribution is to enable machine-aided knowledge extraction for social scientists as well as the general readers of social science research. While the recent years have seen some experiments of NLP models to extract causalities in scholarly papers such as Catalyst Team (2016) and Valenzuela-Escárcega et al. (2018), they were developed on the biomedical literature that often adheres to strict and standard languages and formatting. Recent years have witnessed rising interest to engage NLP techniques in social and business literatures, notably text mining business intelligence, where the main idea is to collect, treat, and store data in order to support decisions of managers (Ishikiriyama, Miro, & Gomes, 2015) and/or to identify emerging topics in the literature. For instance, Moro, Cortez, and Rita (2015) used business research articles related to the banking industries in order to classify them in relevant groups. Ngai, Hu, Wong, Chen, and Sun (2011) also used business articles related to banking industries and found that topics like mortgage fraud, money laundering, and securities and commodities were under-studied. None of these efforts, however, have enabled the extraction and organizing of causalities from the literature, without which the contributions for a "what if" understanding and thus implications for evidence-based practices remain limited.

## 2      Data

### 2.1      Constructing a training sample

We started with developing a training data of 138 empirical articles in social sciences, which were used in a systematic review on organizational performance by Chen, Zhong, Duran, and Sauerwald (2020). These articles were carefully selected from the Web of Science database to represent peer-reviewed and empirically tested papers in social sciences that explicitly discussed issues concerning organizational performance, broadly defined as an organization's effectiveness in meeting expectations of at least two stakeholder groups (investors, employees, customers, and communities). These papers represent a broad literature in social sciences that seek to explain different dimensions of organizational outcomes.

First, using standard tools in Phyton, we converted each PDF to raw text. Due to poor scan quality for optical character recognition (OCR), we lost 13 articles, reducing our dataset to 125 documents. After removing common phrases, tables, figure, and other unnecessary texts from the articles using Phyton packages, we then developed a hypothesis extraction algorithm to identify which statements are likely hypotheses. Specifically, the algorithm works as the following. It detected any statements in a format similar to the following:

***Hypothesis 1** : x and y are positively related.*
***H1** : x and y are positively related.*

We exploited this formatting and trained the algorithm to search for sentences that included targeted expressions like "hypothesis" and "H" (followed by a number or a letter). This gave us 2,230 sentences that potentially contained hypotheses. We ended up with a lot of false positive extractions (i.e., sentences that contained the targeted expressions related to hypotheses, but were in fact not hypothesis statements). We screened all the 2,230 sentences manually and kept true hypothesis statements. We ended up with 643 hypothesis statements across our 125 documents. Below is an example of extracted hypothesis sentences:

*H1. Commitment configuration is positively associated with firm performance.*

Finally, we constructed a strongly balanced training sample of 1,300 sentences by randomly drawing 657 non-hypothesis statements from our training dataset. This training sample in turn was used to train a model for identifying hypothesis statements.

## 2.2 Coding key features of a hypothesis statement

The next task was to develop models to extract information from each hypothesis statement. The objective was to reduce each hypothesis to its four key features like below:

| Node 1 | Node 2 | Direction | Causality |
|---|---|---|---|
| Cause | Effect | +, -, or nonlinear | Causal or Associative but not causal |

We manually classified the nodes and the direction of each hypothesis sentence in our training sample (excluding non-hypothesis sentences). Using the aforementioned hypothesis sentence as an example, the extracted features are:

| Node 1 | Node 2 | Direction | Causality |
|---|---|---|---|
| commitment configuration | firm performance | + | Associative but not causal |

From there, we could train NLP models to classify and extract causes and effects.

## 2.3 A descriptive summary

The 643 hypothesis statements reported a mean of seven hypotheses per article and a standard deviation of 5 hypotheses. As Figure 1 illustrates, the distribution of number of words in each hypothesis statement was approximately normally distribution (with somewhat left skewness) with a mean 18.5 and a standard deviation 9.8, after censoring observations and drooping sentences with more than 70 words. On average a sentence has between 15 and 20 words (Plain English Campaign, 2004).

**Figure 1: Distribution of Number of words in every sentence**

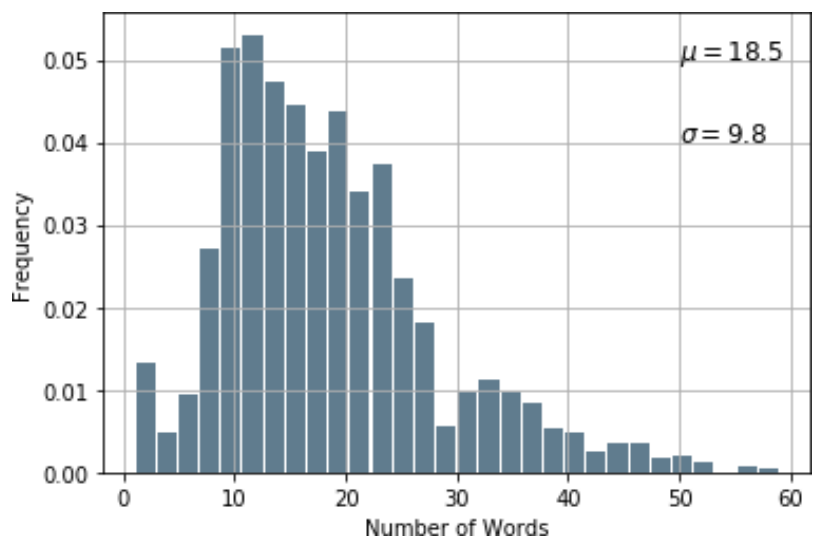

As mentioned, every hypothesis was decomposed into the node of cause, the node of effect, the direction (positive, negative, or nonlinear), and whether the statement has a causal relationship (or not). Among all the hypothesis statements, 84% were classified to have a positive direction between Node 1 and Node 2, and 16% as negative. There were no nonlinear directions. 30% were labeled as causal and 70% as non-causally associative.

## 3 Model training

The following subsections discuss the models we developed to complete the tasks of hypothesis classification, causality classification, and entity extraction.

## 3.1 Hypothesis classification

After constructing the training sample, we trained a model to classify sentences as a *hypothesis sentence* or not. We first manually labeled each sentence as a **hypothesis statement** or not. As mentioned earlier, our training set contained 643 hypothesis statements and, in order to have a relatively balanced final dataset to avoid overfitting, we further randomly chose 657 non-hypothesis sentences from the same sample of articles. Hence, our final training set contained 1,300 sentences.

Unlike a traditional NLP pipeline, text classification models do not seek to understand the meanings or grammatical structure within texts. Instead, we fed the models with features of the training sample in order to find statistical relationships between the raw sentences (inputs) and their labels of features (outputs). As an example, if the word "association" was more related to hypothesis sentences, the model would be more likely to classify sentences with the word "association" as a *hypothesis*, without knowing the meaning of the word "association". Our training sample met the requirement for successful trainings of text classification models, as it covered a wide range of possible hypothesis-related words.

We used the *fastText* – a machine learning (ML) algorithm – as the text classification algorithm to classify sentences as a hypothesis or not. This algorithm was created by Facebook's AI Research group to learn word embeddings and perform text classification; this model has been shown to have similar performance of accuracy as deep learning models but at a faster speed (Zolotov & Kung, 2017).

The algorithm works as the following:

1) It breaks a text apart into separate tokens. Each token is a term or a word;
2) It assigns every word in the training sample an *n*-dimensional numerical vector (word embedding);
3) It assigns every sentence an *n*-dimensional numerical vector that averages the values of every dimension of the word's vectors in the sentence (sentence embedding);
4) The sentence embeddings are finally used as inputs into a linear classification model to predict the final classification label.

During training, our model used a simple neural network with one layer and iterated through word and sentence embeddings so that the embedding for a given sentence and the vector of its associated label were very close to each other in the space. Finally, sentence embeddings were used as input to a linear classification model for final prediction.

We trained several models, whose reports are reported below:

**Table 1: *fastText* - Estimations**

| Word N-grams (1) | Learning Rate (2) | Vector Dimension (3) | F1-Score Soft Max (4) | F1-Score Neg-Sample (5) |
|---|---|---|---|---|
| Model 1 | 1 | 0.1 | 120 | 87.1% | 92.6% |
| Model 2 | 2 | 0.1 | 120 | 84.6% | 85.7% |
| Model 3 | 5 | 0.1 | 120 | 85.1% | 55.3% |
| Model 4 | 1 | 0.3 | 120 | 95.7% | 96.7% |

F1-Score was calculated using two loss functions: Soft Max and Negative-Sampling.

Table 1 presents the best four models after running several experiments. The order of words played no effects on the results of identifying hypothesis sentences. Furthermore, models using bi-grams, compared to those using uni-grams, reported a lower accuracy under all specifications. Also, the negative-sampling loss provided a better accuracy under most specifications. The best specification was Model 4, in which we used uni-grams, a learning rate of 0.3, a 120-dimension vector to represent words, and the negative-sampling loss function. As presented in column (5), we achieved a 96.7% accuracy for this specification on the testing data.

### 3.1.1. LIME - Interpretation of the results

One challenge in ML is the interpretation of results since ML models are often uninterpretable like a "black box". As a result, it cannot be trusted generally that these models have picked up on correct features in the data. For instance, if hypothesis sentences are on average shorter than non-hypothesis sentences in our sample, then the model might have classified short sentences as hypotheses and long sentences as non-hypotheses. In this case, the model would report a high accuracy, but is not working in the correct way we want it to be. Therefore, having an accurate model is sufficient.

Ribeiro, Singh, and Guestrin (2016) introduced an approach to interpreting complex ML models, named Local Interpretable Model-Agnostic Explanations (LIME). Following LIME, we need to explain how the *fastText* model makes a prediction by training a simpler model that mimics it, then use this simpler stand-in model to explain the original *fastText* model's prediction. Even though the simpler model cannot capture all the complexity of the *fastText* model, it helps to understand the logic used to make a single prediction by iterating over the already trained *fastText* model.

Instead of training the stand-in model on the entire training dataset, we used a subsample of the data for the stand-in model to correctly classify one sentence. As long as the stand-in model used the same logic as the *fastText* model, we would be able to understand and explain the predictions made by *fastText*.

In order to construct the training set for the stand-in model, we created many variations out of each sentence, each time removing certain words. In the case of hypothesis classification, we classified a hypothesis sentence many times with different words removed from the sentence in order to understand the importance of each word in the final prediction. By making several predictions for many slight variants of the same sentence using *fastText*, we were essentially capturing how the model understood that sentence. Finally, we used the sentence variations and classifier prediction results as the training set to train the stand-in model using the Simple Linear Classification Model.

We want to note that a shortcoming of this approach is the implicit consideration of only the importance of single words, not phrases or *n*-grams. However, as we will show, this limitation does not prevent us from making reasonable interpretations of *fastText*.

The output of our stand-in model was the weights assigned to each word in the hypothesis sentence, where the weights represent how much that word affected the final prediction.

As shown in Figure 2, the words "positively" and "associated" were among the most important words as they contributed the most to the classification of this sentence as a hypothesis.

**Figure 2: An Example of Hypothesis Sentence**

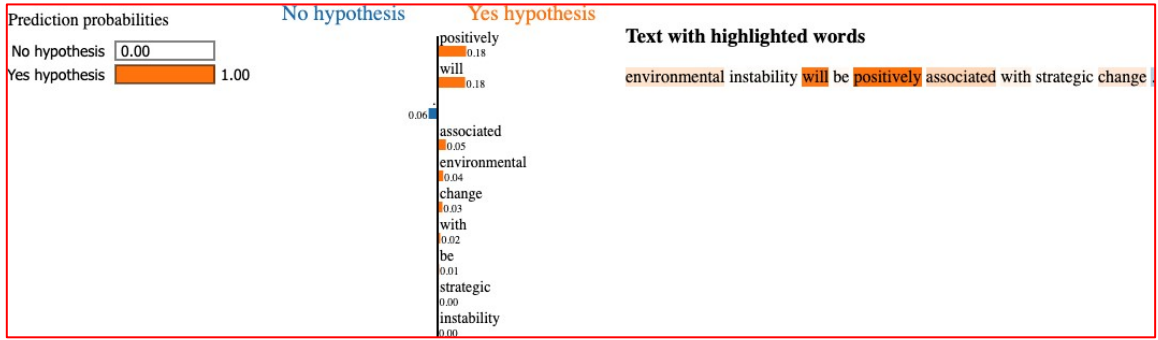

**Figure 3: An Example for a Non-Hypothesis Sentence**

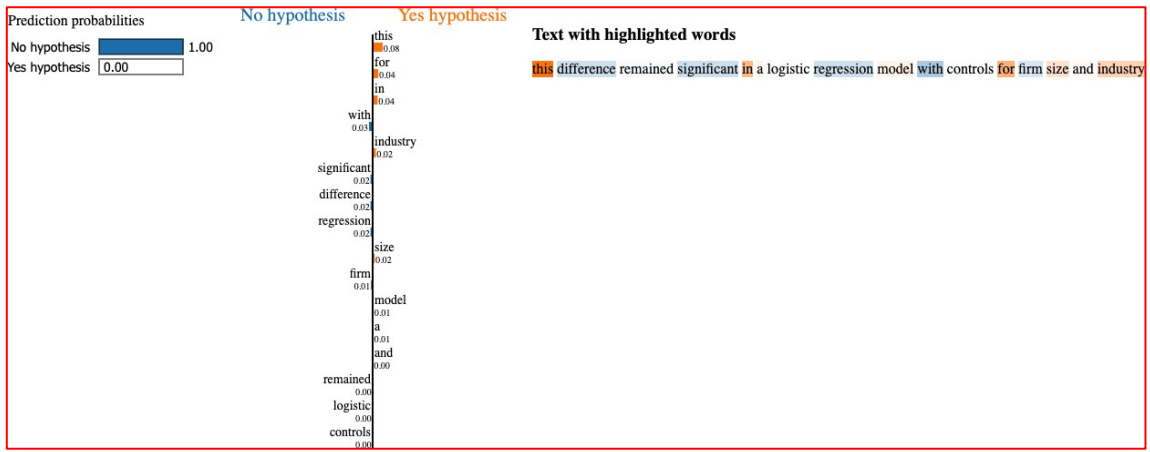

As Figure 3 shows, the words contributing the most to classifying a sentence as a *non-hypothesis* were "significant" and "regression". They are usually not part of a hypothesis sentence. However, no word in this sentence was strongly associated with a hypothesis sentence. Therefore, from these two figures, it seems clear that the *fastText* model was valuing the correct words to make predictions regarding hypothesis classification.

### 3.2   Causality classification

We then moved on to train a model to classify if a sentence made a causal statement or not. In addition to basic text cleaning (e.g., tokenization, normalization, and removal of stop words), we replaced the cause- and effect entities with "node1" and "node2." For example, suppose we have the sentence below:

*Playing music causes better concentration.*

After cleaning we would have the sentence:

*Node1 causes better node2.*

The reason for doing this is to reduce the amount of unnecessary information in the data. In this task, it is not about what "node1" and "node2" are, but about identifying if the relationship between them is causal or not.

Next, we created two different features from the sentences. The first feature was a bag-of-words (BOW) feature for which we identified the frequency of tri-grams against the whole corpus. The second feature was a sentence embedding using Doc2Vec (D2V). We compared models using the BOW features and the D2V features. Tables 2 and 3 report the results for the models on the testing data using the BOW feature and using the D2V feature respectively. As reported in Tables 2 and 3, we found that models using BOW features performed better on the testing data, producing an F1-score greater than 90%. We trained our models using logistic regression and SVM (linear kernel).

**Table 2: Accuracy - BOW**

|  | F1-Score |
|---|---|
| Logistic Regression | 94% |
| SVM | 94% |

**Table 3: Accuracy - D2V**

|  | F1-Score |
|---|---|
| Logistic Regression | 64% |
| SVM | 69% |

### 3.3 Entity extraction

Lastly, we trained our model to extract two entities in a sentence that are related. For example, if we have:

*Node1 is related to node2.*

Then, we want to extract both "node1" and "node2." But if we have a causal statement:

*Node1 causes node2.*

Then we need to not only extract "node1" and "node2," but also identify "node1" as the causal entity and "node2" as the effect entity.

**Figure 4: Entity Extraction Model Performance**

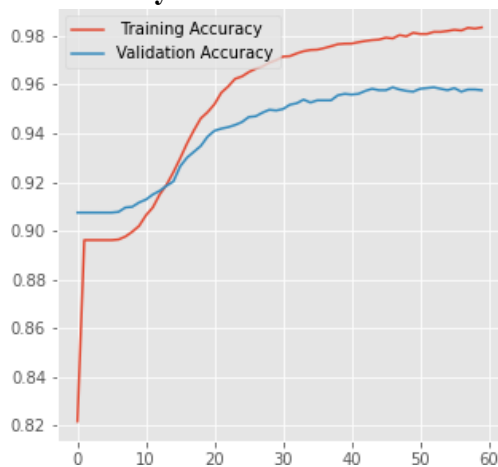

First, we reconstructed our sample data. For each sentence, we label non-entities as 0, the "cause" entity as "1", and the "effect" entity as "2". We padded each of the sentences so they were formatted to have a dimension of 70 (i.e., the vector dimension). We then fitted the data to a bi-directional LSTM with three units and, after many experiments, run the model with a batch size of 32 and 60 epochs. Figure 4 shows the training and validation accuracy over number of experiments.

Furthermore, from our testing data, we had an overall accuracy of 95%. However, when just looking at the sub task of identifying the "cause" entity and the "effect" entity, we saw an accuracy of 80% and an accuracy 70%, respectively. This may be because the cause and effect entities are sparse compared to the non-entities, and the model is very good at classifying the non-entities. However, the model performed very well when identifying the cause and effect entities.

## 4   Discussion and conclusion

In this project, we first constructed our own training set of hypothesis statements from a set of scientific papers in social sciences. Then we used this training set to train models that perform three different tasks. In our first task, we used the *fastText* model to predict whether a sentence is a hypothesis statement or not. Using negative sampling, the highest F1-score we obtained was 96.7%. *fastText* model was capturing the salient features to distinguish between hypothesis and non-hypothesis statements. Using the LIME model, we found that *fastText* was classifying our sentences in a way that was interpretable. The second task was causality classification. For this task, we found that logistic regression model had the best performance. Specifically, we obtained an F1-Score of 94%. The final task was entity extraction. We used a bi-directional LSTM to decompose the causal statements into cause- and effect entities, generating an overall accuracy of about 95%. The accuracy for the sub-task of the entity identification was also satisfactory, reporting an accuracy of 80% and 70% respectively for identifying the cause entities and effect entities separately.

We suggest a few valuable extensions in future studies. First, our corpus was entirely based on scientific papers, which represent a limited set of textual resources both researchers and practitioners use to draw knowledge from social sciences. We suggest it is valuable to extent the three tasks in other valuable documents such as teaching cases, textbooks, industry reports, policy documents, and news articles. Second, another value extension is to accelerate the construction of the training dataset. In our work, we manually constructed our training data. However, this task may become exceedingly difficult as the size of the corpus increases. Therefore, a valuable extension is to automate the construction of training data.